\pdfoutput=1
\documentclass[11pt]{article}

\usepackage[]{ACL2023}

\usepackage{times}
\usepackage{latexsym}
\usepackage{graphicx}

\usepackage[T1]{fontenc}
\usepackage[utf8]{inputenc}

\usepackage{microtype}

\usepackage{inconsolata}

\title{Rapid Speaker Adaptation in Low Resource Text to Speech Systems using Synthetic Data and Transfer learning}

\author{Raviraj Joshi \\
  Flipkart, Bengaluru \\
  \texttt{raviraj.j@flipkart.com} \\\And
  Nikesh Garera \\
  Flipkart, Bengaluru \\
  \texttt{nikesh.garera@flipkart.com} \\}

\begin{document}
\maketitle
\begin{abstract}
Text-to-speech (TTS) systems are being built using end-to-end deep learning approaches. However, these systems require huge amounts of training data. We present our approach to built production quality TTS and perform speaker adaptation in extremely low resource settings. We propose a transfer learning approach using high-resource language data and synthetically generated data. We transfer the learnings from the out-domain high-resource English language. Further, we make use of out-of-the-box single-speaker TTS in the target language to generate in-domain synthetic data. We employ a three-step approach to train a high-quality single-speaker TTS system in a low-resource Indian language Hindi. We use a Tacotron2 like setup with a spectrogram prediction network and a waveglow vocoder. The Tacotron2 acoustic model is trained on English data, followed by synthetic Hindi data from the existing TTS system. Finally, the decoder of this model is fine-tuned on only 3 hours of target Hindi speaker data to enable rapid speaker adaptation. We show the importance of this dual pre-training and decoder-only fine-tuning using subjective MOS evaluation. Using transfer learning from high-resource language and synthetic corpus we present a low-cost solution to train a custom TTS model. 
\end{abstract}

\section{Introduction}
\label{sec:intro}
Speech synthesis systems are widely used in applications like voice assistants and customer service voice bots \cite{joshi2021attention1,joshi2022simple}. They are used commonly along with automatic speech recognition (ASR) \cite{joshi2022comparison1} systems to provide an end-to-end voice interface. Recently, text-to-speech (TTS) systems have been trained using end-to-end deep learning approaches \cite{shen2018natural}. The TTS models are based on an independent acoustic model converting text to spectrogram and a vocoder converting spectrogram to speech. More recently, these two models have been integrated into the model directly converting text to target speech \cite{weiss2021wave}. However, the single end-to-end model requires large amounts of transcribed data. The dual model approach can be trained on comparatively less data as training a vocoder only requires audio data without its text transcripts. In general, training an end-to-end TTS requires a large amount of high-quality studio recordings to build a production-quality model. 

\begin{figure}[tb]
  \centering
  \includegraphics[scale=0.3]{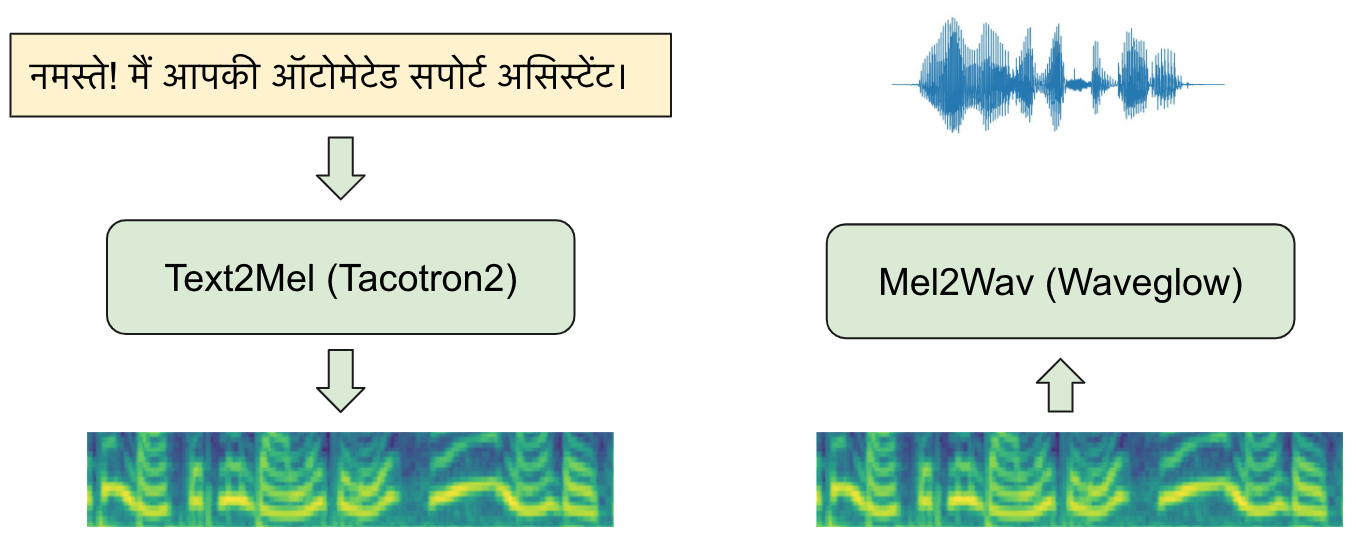}
  \caption{Overall flow of the TTS system}
  \label{fig:tts_model_flow}
%   \Description{Basic LAS architecture used in our work}
\end{figure}

The popular text to spectrogram models include Tacotron2 \cite{shen2018natural}, Transformer-TTS \cite{li2019neural}, FastSpeech2 \cite{ren2020fastspeech}, FastPitch \cite{lancucki2021fastpitch}, and Glow-TTS \cite{kim2020glow}. In terms of voice quality the Tacotron2 model is still competitive with other models and less prone to over-fitting in low resource settings \cite{favaro2021itacotron,abdelali2022natiq,garcia2022evaluation,finkelstein2022training}. There are multiple options for the vocoder as well like Clarinet \cite{ping2018clarinet}, Waveglow \cite{prenger2019waveglow}, MelGAN \cite{kumar2019melgan}, HiFiGAN \cite{kong2020hifi}, StyleMelGAN \cite{mustafa2021stylemelgan}, and ParallelWaveGAN \cite{yamamoto2020parallel}. We choose Waveglow since it is competitive with other vocoders and is easy to train \cite{abdelali2022natiq,garcia2022evaluation,shih2021rad}. 
%In this work, we are not concerned about inference speed and therefore choose Tacotron2 + Waveglow architecture for our data-oriented experiments. 
%In this work, we are not concerned about model architectures and therefore choose the popular Tacotron2 + Waveglow architecture for our data-oriented experiments.
There is other single model end-to-end architectures like VITS \cite{kim2021conditional}, Wave-Tacotron \cite{weiss2021wave} and JETS \cite{lim2022jets} for spectrogram-free TTS approaches. Although such models are more desirable since they remove the vocoder spectrogram features mismatch during training and inference but do not work well in low-resource settings.

In this work, we explore the Tacotron2-based acoustic model and Waveglow-based vocoder to build a production-quality TTS system in low-resource, low-budget settings. The high-level flow is shown in Figure \ref{fig:tts_model_flow}. In general, these models require 10 to 20 hours of quality data to train high-quality TTS systems. We aim to reduce the data requirements using simple strategies. Previous works in literature have proposed approaches to adapt to a new speaker with a few hours to a few minutes of data \cite{prakash2020generic}. However, these approaches have only been tested on some 10-30 utterances and might not be suitable for high-quality applications. Recently, a TTS system Vall-E \cite{wang2023neural} has shown extraordinary zero-shot capabilities. However, this system uses a complex architecture and requires 60K hours of pre-training data making it infeasible in low-resource scenarios.

In order to build low-resource TTS, we explore transfer learning from English data and synthetic audio corpus from the existing TTS model. We show the effectiveness of our approach in the context of the low-resource Indian language Hindi. While transfer learning from English is a common approach, we propose the usage of existing out-of-the-box TTS to further augment the data. Using an out-of-the-box TTS system has multiple advantages. It allows us to get a large amount of (real-text, synthetic-audio) pairs in the domain of our choice. It is a low-cost solution as compared to obtaining studio recordings for an equivalent amount of data. With high-quality out-of-the-box single-speaker text-to-speech systems available in the majority of languages, we leverage it to build a TTS in the speaker of our choice. We use an in-house single-speaker Hindi TTS system to generate synthetic corpus, however, the approach is applicable to any out-of-the-box TTS system. While we could have directly used the original training data of the initial TTS system, we utilized the model as a black box so that we can generate data in the domain of our choice. 

\begin{figure}[tb]
  \centering
  \includegraphics[scale=0.25]{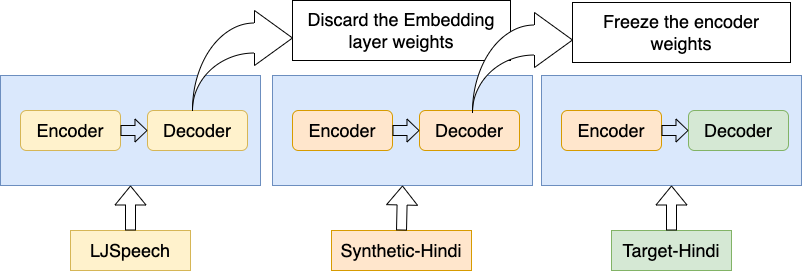}
  \caption{The training strategy for low resource TTS using less amount of Target-Hindi Data}
  \label{fig:tts_flow}
%   \Description{Basic LAS architecture used in our work}
\end{figure}

We propose a three-step approach to build a low-cost TTS system. This approach is depicted in Figure \ref{fig:tts_flow}. 
\begin{itemize}
    \item We initially pre-train the Tacotron2 acoustic model with public English LJSpeech data.
    \item We then ignore the initial character embedding layer based on English (Roman script) and re-train the entire model using a synthetic Hindi corpus (Devanagari script). This synthetic data pre-training step is important as it adapts the Tacotron2 encoder to the target domain text in the Devanagari script. 
    \item Finally, we adapt the model to the target speaker using 3 hours of target speaker Hindi data. In the second step, although the audios are synthetic and from a different speaker, a large amount of real target domain text ensures high-quality pretraining of the text encoder. Therefore, during the final step, we freeze the encoder and only fine-tune the decoder of the Tacotron2 encoder-decoder model.
\end{itemize}
  We show that using this three-step strategy allows us to rapidly build a TTS system in the speaker of our choice. Although we can further reduce the data requirements overall stability of the model is impacted thus hindering its deployment in high-quality applications. We perform a subjective evaluation of our approach on an unseen domain with a larger test set and show its effectiveness.      

\section{Related Work}
\label{sec:format}
In this section, we describe the previous attempts to train a low-resource TTS system. 
A generic Indic TTS system using multiple languages and voices was built in \cite{prakash2020generic}. They used the tacotron2 + waveglow setup, along with a common linguistic representation multi-language character map (MLCM) to map all languages to a common script. They also utilize speaker embeddings based on x-vectors to train a multi-speaker model. Their observations indicate that such a system does not scale to the unseen speakers. So they propose a new speaker adaptation using only 7 minutes of data. However, the adapted system was tested only on 10 utterances which might not scale well to a larger set in high-quality settings. In this work, we focus on the original script of the language as mapping it to a common script could result in information loss for some languages.  Previously, a multi-lingual TTS system using MLCM character representation was introduced in \cite{prakash2019building}. 

Cross-lingual transfer learning and data augmentation approach for low resource TTS were proposed in \cite{byambadorj2021text}. The spectrogram prediction network was trained using cross-lingual transfer learning (TL) from high resource language, data augmentation by varying parameters like pitch and speed, and a combination of two approaches. In the TL approach models were sequentially trained on high resource language (English) followed by a low-resource language (Mongolian). The input script was first converted into IPA phonetic format to enable the transfer of knowledge. We followed a similar approach in our first two steps but without using the common IPA phones instead of relying on the original script. This work also indicates a minimum of 3 hours of data is needed to cover all the phones. However, this approach is based on a multi-speaker TTS system as opposed to our single-speaker model. Similarly, data augmentation using a voice conversion module was explored in \cite{cai2023cross} and \cite{ribeiro2022cross}.

Another approach for data augmentation using the parent TTS system was proposed in \cite{hwang2021tts}. An auto-regressive TTS was first trained and used to generate large-scale synthetic corpora. This synthetic corpus along with real corpus is then used to train a non-auto-regressive TTS system. A similar approach utilizing synthetic corpus from existing TTS is explored in \cite{finkelstein2022training,song2022tts}. Our work is similar to these approaches where the common aspect is to generate synthetic audio from another TTS system. However, these works train different TTS architectures for the same speaker and propose complicated training approaches. We instead focus on any-speaker out-of-the-box TTS system and propose a simple fine-tuning strategy.

Multi-speaker models leveraging external speaker embedding are commonly used to address speaker adaptation in low-resource settings. Transfer learning from external speaker verification models to Tacotron2 TTS was initially explored in \cite{jia2018transfer}. Further, zero-shot un-seen speaker adaptation using a similar speaker embedding approach was explored in \cite{cooper2020zero}. Other approaches have been proposed over to time to combine speaker embedding and style embedding in Tacotron2 setup \cite{chung2021fly}. In this work, we are only concerned about single-speaker models in this work and directly use the input script to preserve the original representation. 

\section{Model Architecture}
\label{sec:format}

We use a Tacotron2-based spectrogram prediction network followed by a Waveglow-based speech synthesis model. The two pass models are preferred in low resource settings as compared to fully end-to-end models. We observe that the vanilla Tacotron2 model is less prone to over-fitting in low-resource scenarios as compared to the vanilla Transformer based Tacotron model. Moreover, these models are competitive with more recent models in terms of audio quality and hence used to evaluate our transfer learning approaches \cite{tan2021survey}. The approaches presented in this work are data-oriented and can be easily extended to any other model architecture.
\subsection{Tacotron2}
The Tacotron2 \cite{shen2018natural} model uses an auto-regressive spectrogram prediction network followed by a wavenet vocoder \cite{vanwavenet}. We describe the details of the spectrogram prediction network from Tacotron2 used in this work. The text-to-Mel spectrogram prediction is done using a sequence-to-sequence network. The input characters are converted into 512-dimensional embeddings and passed to the encoder-decoder network. The encoder consists of 3 stacked convolution layers with 512 filters and a filter size of 5 x 1. Each convolution layer is followed by batch normalization and relu activation. The convolution block is followed by a single Bi-LSTM layer with 512 units. The decoder is an auto-regressive network conditioned on encoder hidden representation. It uses location-sensitive attention \cite{chorowski2015attention} to compute the context vector during each time step. The decoder uses pre-net, containing 2 feed-forward layers (256 units) followed by relu units. The output of the pre-net is concatenated with the context vector and passed through two uni-LSTM layers with 1024 units. The output of lstm is again concatenated with the context vector and passed through a dense layer to predict the spectrogram frame. At this step, another parallel projection predicts the stop token. The predicted frame is passed through a post-net comprising of 5 conv layers (512 filters, 5 x 1 filter size) each followed by batch norm and tanh non-linearity. The post-net predicts the residual to be added to spectrogram prediction to enhance the output. The mean squared error (MSE) is used as a loss function. The loss is computed on both output of LSTM projection and post-net projection. 

The ground truth mel spectrogram is computed with STFT using 50 ms frame length and 12 ms hop. The STFT magnitude is transformed into a Mel scale using an 80-channel Mel filter bank. This is followed by log compression to get ground truth log-Mel spectrogram. Other hyperparameters are the same as those described in the original work.

\subsection{Waveglow}
Waveglow \cite{prenger2019waveglow} is a flow-based network that converts Mel-spectrogram to speech. It is a generative model that generates audio by sampling from a distribution. The samples are taken from zero mean, spherical Gaussian distribution, and transformed into audio distribution by passing it through a series of invertible transformations. It essentially models the audio distribution conditioned on a Mel-spectrogram. The model is trained by minimizing the log-likelihood of the data.    
\section{Dataset Details}
\label{sec:dataset}

We use three different datasets in this work. These datasets are single-speaker labeled audio-text pairs. One dataset is synthetically generated while the other two are real data. These are described below.  
\begin{itemize}
    \item \textbf{LJSpeech-English (24 hrs)}: It is a public domain single-speaker audio dataset consisting of 13,100 audio clips \cite{ljspeech17}. The text of the audio is taken from 7 non-fiction English books. The total length of the dataset is approximately 24 hours.
    \item \textbf{Syntethic-Hindi (15 hrs)}: A synthetic data set is created using an in-house TTS system. The output of the system was a single speaker and a female voice. Around 16k short utterances in Devanagari script majorly from the grocery voice assistant domain were converted to speech. The size of this dataset is around 15 hrs. 
    \item \textbf{Real-Hindi (3 hrs)}: This is the target low-resource speaker data. A subset of utterances (2.5k) from the voice assistant domain were recorded in the voice of an external female speaker. These were high-quality studio recordings and the size of the dataset was around 3 hrs. We also evaluate the full 15 hrs of studio recordings of the above voice assistant utterances for comparative analysis.
\end{itemize}
All the audio data is re-sampled at 16kHz and encoded in 16-bit PCM wav format for training and inference.

\begin{table*}
  \centering
  \begin{tabular}{cc}
    % \toprule
    \hline
    \textbf{Training Strategy} & \textbf{MOS}\\
    % \midrule
    \hline
    Ground Truth Audios (Real) & 4.65 $\pm$ 0.62  \\ \hline
    LJSpeech + Real (3 hrs) & 4.27 $\pm$ 0.95 \\ \hline
    LJSpeech + Synthetic + Real (3 hrs) & 4.54 $\pm$ 0.58\\ \hline
    LJSpeech + Synthetic + Real (frozen encoder, 3 hrs) & \textbf{4.59} $\pm$ 0.68  \\ \hline
    LJSpeech + Synthetic + Real (frozen encoder, 15 hrs)& 4.65 $\pm$ 0.58 \\ \hline
%   \bottomrule
\end{tabular}
\caption{MOS scores for different training strategies}
  \label{tab:all_res}
\end{table*}

\section{Experimental Setup}
\label{sec:exp_setup}
We evaluate different variations of the pre-training strategy and try to come up with the best training strategy. We use English data and synthetic data for pre-training. First, the model is trained on English data followed by Hindi synthetic data. Finally, the model is trained on the target real Hindi corpus. The third corpus is the smallest in size and depicts the low-resource speaker. The final fine-tuning is done in two different ways. One approach is to perform full-finetuning and the second approach is to perform partial fine-tuning by freezing the encoder. The frozen encoder approach yields the best results and the corresponding flow is shown in Figure \ref{fig:tts_flow}. The frozen text encoder is desirable since it is pre-trained on a large amount of target domain text as opposed to a small amount of data in the third step.
Overall we consider the following pre-training strategies:
\begin{itemize}
    \item \textbf{Direct target speaker (Hindi) training} - With just 3 hours of data and the results were mostly noisy and the training did not converge to a decent model. So the results of this model training are not discussed in the next sections.
    \item \textbf{LJSpeech English pre-training + Target-Hindi finetuning} - In this setup, since the input symbols of English and Hindi are completely different we discard the embedding layer weights and retain all other weights of the pre-trained English model. Post this we perform target Hindi data fine-tuning. Also, full fine-tuning is performed using target data since the embedding layer is part of the encoder and needs to be re-trained.
    \item \textbf{LJSpeech English pre-training + Synthetic Hindi pre-training + Target-Hindi full finetuning} - In this strategy, the model is initially trained on English corpus, followed by full training on synthetic Hindi corpus. Finally, target Hindi speaker data is used to again fully fine-tune the models.
    \item \textbf{LJSpeech English pre-training + Synthetic Hindi pre-training + Target-Hindi decoder only finetuning} - This strategy is similar to the last strategy. The only difference is in the final fine-tuning only decoder weights are updated and the encoder is completely frozen. With this, we try to retain learnings from a much larger Hindi corpus. 
\end{itemize}

\begin{figure}[tb]
  \centering
  \includegraphics[scale=0.25]{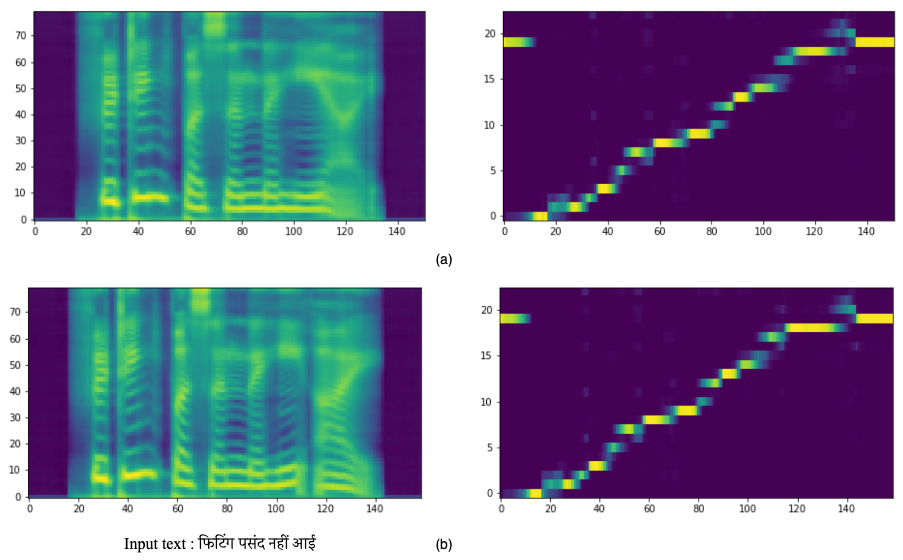}
  \caption{The Mel-spectrogram and attention alignment plot for a sample sentence using config (a) LJSpeech + Real (3 hrs) and (b) LJSpeech + Synthetic + Real (frozen encoder, 3 hrs). The difference in the resolution can be clearly seen at the end of the two spectrograms. }
  \label{fig:tts_spectrogram}
%   \Description{Basic LAS architecture used in our work}
\end{figure}

\section{Results and Discussion}
\label{sec:results}

We evaluate different pre-training strategies explored in this work using subjective mean opinion score (MOS) evaluation. A test set was created using 200 consumer experience (CX) voice bot interaction utterances. The domain of the test data (CX) is different from the domain of training utterances (voice assistant). This allows us to perform more rigorous testing of the model to unseen domains. These utterances were evaluated on (1-5) MOS scale by 10 specialized listeners with each audio evaluated by at least 3 listeners. The evaluators were explicitly trained for the evaluation activity and were part of the internal operations team thus ensuring high quality of evaluation. The audio were rated based on intelligibility and naturalness of the audios.

The results of the evaluation are shown in Table \ref{tab:all_res}. The results indicate that LJSpeech English pre-training is helpful to create usable TTS models. Without this cross-lingual transfer learning, the model fails to produce intelligible output. Further training the model on synthetic Hindi data improves the output even further. Synthetic data pre-training is evaluated in two configurations. The full model is fine-tuned in the first config and the encoder is frozen in the second config. In both configurations, we perform LJSpeech pre-training. We observe that the frozen encoder approach yields superior performance in low-resource settings. All these experiments used 3 hrs of real Hindi data, 15 hours of synthetic Hindi data, and 24 hours of real English corpus. We also evaluate the three-step approach using full 15 hrs real data and see minimal improvements in performance. This indicates that 3 hrs of data is sufficient to build a production-ready TTS system with transfer learning from cross-lingual data and synthetic corpus. The plot of spectrogram and attention alignment weights for a sample sentence with and without using synthetic Hindi data is shown in Figure \ref{fig:tts_spectrogram}.

\section{Conclusion}
\label{sec:conc}

We present our transfer learning strategy to build low resource TTS system. We explore transfer learning from cross-lingual data and same-language synthetic data. The synthetic data is created using existing out of the box TTS system. The three-step approach involves pre-training with English data followed by synthetic Hindi data and low-resource real Hindi data. We evaluate these pre-training approaches using a strong out-of-domain test set using subjective MOS evaluation. In the final step, we observe that the decoder only fine-tuning works better than full tuning. The high-level text representations (encoder output) of text trained on the large real text and synthetic audio pairs are better than just using the low-resource data.

\bibliography{main}
\bibliographystyle{acl_natbib}

% \appendix

% \section{Example Appendix}
% \label{sec:appendix}

% This is a section in the appendix.

\end{document}